\documentclass[runningheads]{llncs}
\usepackage{graphicx}
\usepackage{comment}
\usepackage{amssymb} 
\usepackage{todonotes} 
\usepackage{caption}
\usepackage{subcaption}
\usepackage{tabularx}
\usepackage[hidelinks]{hyperref}

\begin{document}
\title{Ontology Matching Through Absolute Orientation of Embedding Spaces}

\author{Jan Portisch\inst{1,2}\orcidID{0000-1111-2222-3333} \and
Guilherme Costa\inst{1}\orcidID{0000-0001-6188-9868 } \and
Karolin Stefani\inst{1}\orcidID{0000-0003-0113-6842} \and
Katharina Kreplin\inst{1}\orcidID{0000-0003-3467-8664} \and
Michael Hladik\inst{1}\orcidID{0000-0001-5420-0663} \and
Heiko Paulheim\inst{2}\orcidID{0000-0003-4386-8195}
}
\authorrunning{J. Portisch et al.}
%
\institute{SAP SE, Walldorf, Germany\\
\email{\{jan.portisch, guilherme.costa, karolin.stefani, katharina.kreplin, michael.hladik\}@sap.com}\\
\and
Data and Web Science Group, University of Mannheim, Germany\\
\email{\{jan,heiko\}@informatik.uni-mannheim.de}}

\maketitle              
\begin{abstract}
Ontology matching is a core task when creating interoperable and linked open datasets. 
In this paper, we explore a novel structure-based mapping approach which is based on knowledge graph embeddings: The ontologies to be matched are embedded, and an approach known as absolute orientation is used to align the two embedding spaces.
Next to the approach, the paper presents a first, preliminary evaluation using synthetic  and real-world datasets. We find in experiments with synthetic data, that the approach works very well on similarly structured graphs; it handles alignment noise better than size and structural differences in the ontologies. 

\end{abstract}
\section{Introduction}
Ontology matching describes the complex process of finding an alignment $A$ between two ontologies $O_1$ and $O_2$. An alignment is a set of correspondences where a correspondence is, in its simplest form, a tuple of $\langle e_1, e_2, R \rangle$ where $e_1 \in O_1$ is an element from one ontology, $e_2 \in O_2$ is an element from the other ontology, and $R$ is the relation that holds between the two elements. Typically, $R$ is equivalence ($\equiv$).

In this paper, we examine the use of embedding two ontologies for finding an alignment between them. Given two embeddings of the ontologies, we use a set of anchor points to derive a joint embedding space via a rotation operation.

\section{Related Work}

\subsubsection{Knowledge Graph Embeddings}
Given be a (knowledge) graph $G = (V,E)$ where $V$ is the set of vertices and $E$ is the set of directed edges. Further given be a set of relations $R$, $E \subseteq VxRxV$. A knowledge graph embedding is a projection $E \cup R \rightarrow \mathbb{R}^d$.\footnote{Variations of this formulations are possible, e.g., including different dimensions for the vector spaces of $E$ and $R$, and/or using complex instead of real numbers.} In this paper, we use the RDF2vec approach, which generates multiple random walks per vertex $v \in V$. An RDF2vec sentence resembles a walk through the graph starting at a specified vertex $v$. 
Those random walks are fed into a \textit{word2vec} algorithm, which treats the entities and relations as words and the random walks as sentences, and consequently outputs numeric vectors for entities and relations.

\subsubsection{Absolute Orientation} 
Multiple approaches exist for aligning embeddings. In this paper, the extension by Dev et al.~\cite{DBLP:journals/kais/DevHP21} of the \emph{absolute orientation} approach is used. The approach showed good performance on multilingual word embeddings. The calculation of the rotation matrix is based on two vector sets $A = \{a_1, a_2, ... a_n\}$ and $B = \{b_1, b_2, ... b_n\}$ of the same size $n$ where $a_i, b_i \in \mathbb{R}^d$.
In a first step, the means $\bar{a} = \frac{1}{n} \sum^{n}_{i=1} a_i$ and $\bar{b} = \frac{1}{n} \sum^{n}_{i=1} b_i$ are calculated. Now, $\bar{a}$ and $\bar{b}$ can be used to center $A$ and $B$: $\hat{A} \leftarrow (A, \bar{a})$ and $\hat{B} \leftarrow (B, \bar{b})$.
Given the sum of the outer products $H = \sum^{n}_{i=1}\hat{b}_{i}\hat{a}^{T}_i$, the singular value decomposition of H can be calculated: $svd(H) = [U, S, V^T]$. The rotation is $R=UV^T$. Lastly, $\hat{B}$ can be rotated as follows: $\widetilde{B} = \hat{B}R$.

\subsubsection{Matching with Embeddings}
Embedding-based matching approaches have gained traction recently, mostly using embeddings of the textual information contained in ontologies~\cite{portischbackground}. OntoConnect~\cite{DBLP:conf/icmla/ChakrabortyZSB21}, for example, uses fastText within a larger neural network to match ontologies; DOME~\cite{DBLP:conf/semweb/HertlingP19} exploits doc2vec; TOM~\cite{DBLP:conf/semweb/KossackBKP21} and F-TOM~\cite{DBLP:conf/semweb/KnorrP21} use S-BERT. With the exception of ALOD2vec Matcher~\cite{DBLP:conf/semweb/PortischP21a}, knowledge graph embeddings are rarely used.
The work presented in this paper is different in that it does not rely on labels or an external knowledge graph. Instead, an embedding is learnt for the ontologies to be matched. 

\section{Approach}
We first train two separate embedding spaces for the two ontologies to be matched (i.e., $O_1$ and $O_2$). This is done in two independent RDF2vec training processes. In a second step, we then perform the absolute orientation operation to rotate one embedding space onto the other.

For the matching operation, we assign for each node in $e \in O_1$ the closest node $e \in O_2$ according to Euclidean distance.

\begin{figure}
  \centering
  \includegraphics[width=\linewidth]{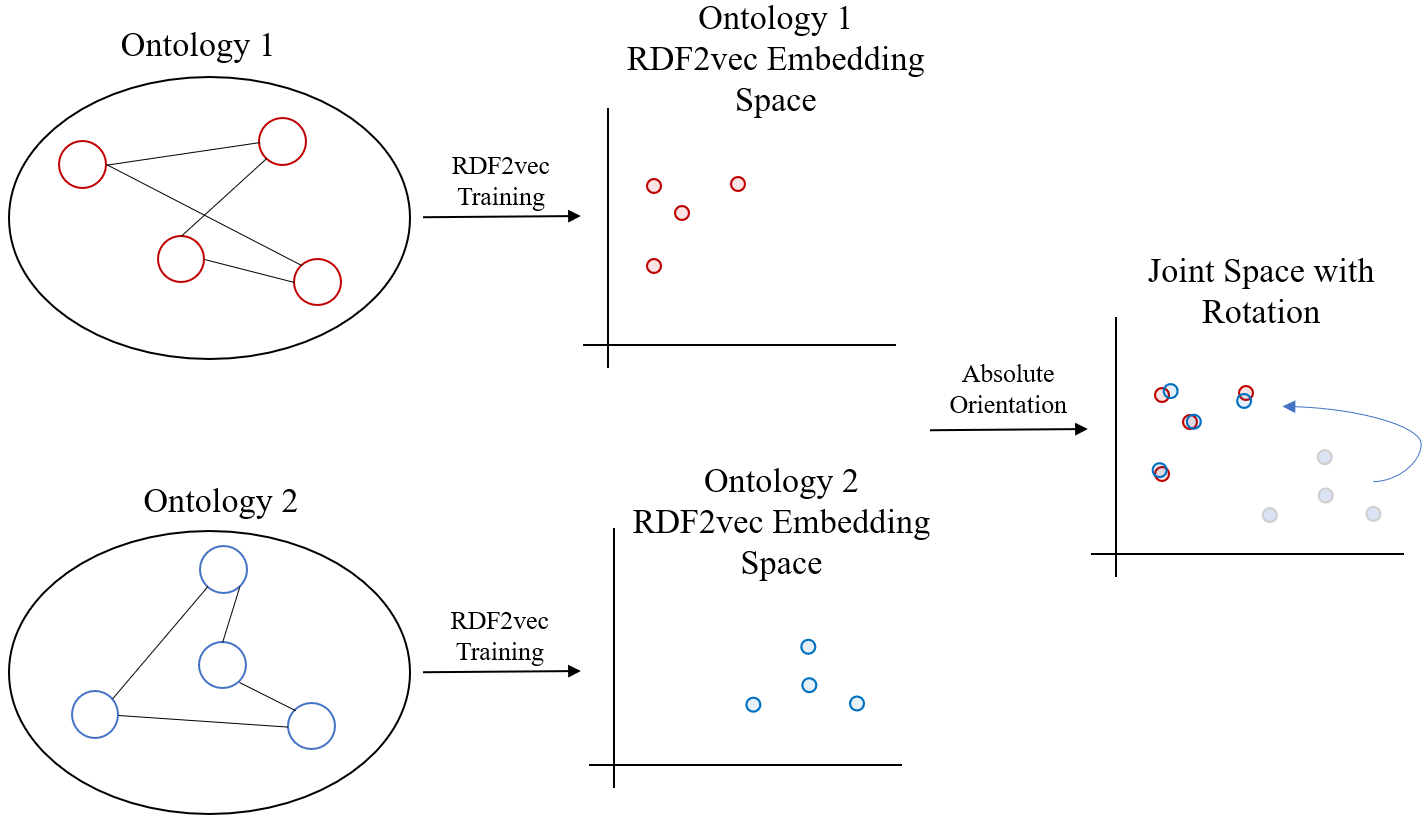}  
  \caption{High-level overview of the absolute orientation approach.}
  \label{fig:overview}
\end{figure}

\section{Experiments}
For the experiments, jRDF2vec\footnote{see \url{https://github.com/dwslab/jRDF2Vec}}~\cite{jrdf2vec} was used to obtain RDF2vec embeddings. We chose the following hyper parameter values: $dimension=100$, $window=6$, $depth=6$, $walks=150$. The code together with the complete set of figures and results is available online.\footnote{see \url{https://github.com/guilhermesfc/ontology-matching-absolute-orientation}}

\subsection{Synthetic Experiments}
In a first step, we perform sandbox experiments on synthetic data. We generate a graph $G$ with 2,500 nodes $V$. For each node $v \in V$, we draw a random $d$ number using a Poisson distribution $f(k;\lambda) = \frac{\lambda^ke^{-\lambda}}{k!}$ with $\lambda=4$. We then randomly draw $d$ nodes from $V \setminus v$ and add the edge between $v$ and the drawn node to $G$.
We duplicate $G$ as $G'$ and generate an alignment $A$ where each $v \in V$ is mapped to its copy $v' \in V'$. We define the matching task such that $G$ and $G'$ shall be matched. The rotation is performed with a fraction $\alpha$ from $A$, referred to as the anchor alignment $A'$. In all experiments, we vary $\alpha$ between $0.2$ and $0.8$ in steps of size $0.2$.

\paragraph{Training Size}
In order to test the stability of the performed rotation, also referred to herein as training, we evaluate varying values for $\alpha$.
Each experiment is repeated 5 times to account for statistical variance.
The matching precision is computed for each experiment on the training dataset $A'$ and on the testing dataset $A \setminus A'$.
The split between the training and the testing datasets is determined by $\alpha$.
We found that the model is able to map the entire graphs regardless of the size of the training set $A'$ (each run achieved a precision of 100\%). 

\begin{figure}
    \centering
    \includegraphics[width=1.0\textwidth]{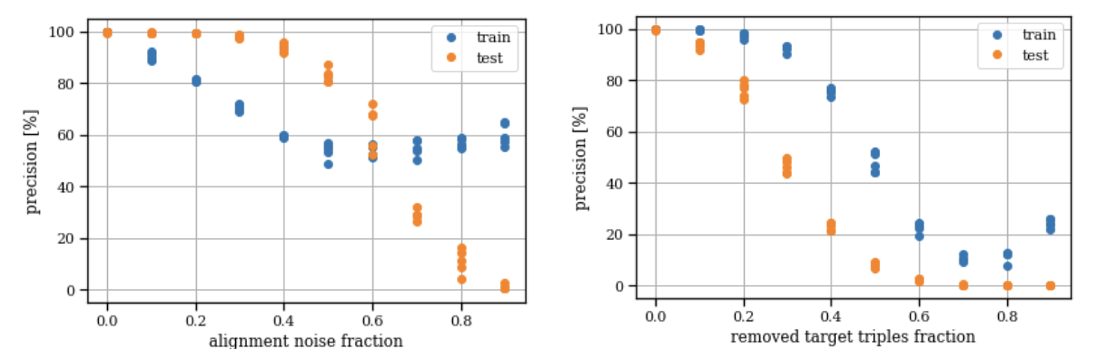}
    \caption{The effect of distortions. (1) alignment noise (left) and (2) size differences (right). Graphs are given for $\alpha=0.2$.}
    \label{fig:plots-02}
\end{figure}

\paragraph{Alignment Noise}
In order to test the stability in terms of noise in the anchor alignment $A'$, we distort a share of the training correspondences by randomly matching other than the correct nodes.
We vary this level of alignment noise between 0 (no noise introduced) and 0.9 ($90\%$ of the alignments are randomly matched) in steps of size $0.1$. Figure~\ref{fig:plots-02} (left) shows the performance with $\alpha=0.2$. 
We observe that the test performance declines with an increasing amount of noise. Interestingly, this relation is not linear. It is visible in Figure~\ref{fig:plots-02} (left) that the approach can handle 40\% of noise before dropping significantly in terms of test performance. 

\paragraph{Graph Heterogeneity}
In order to test the stability in terms of graph heterogeneity, we randomly remove triples from the target graph $G'$ after setting up the alignment between the source graph $G$ and the target graph $G'$. We vary the fraction of randomly removed triples in $G'$ between 0 (no triples removed) and 0.9 ($90\%$ of the triples removed) in steps of size $0.1$. In Figure~\ref{fig:plots-02} (right) it can be observed that with a size deviation of 30\%, the performance starts to drop rapidly. Comparing the two plots in the figure, it can be seen that the approach handles noise significantly better than size and structure deviations in graphs.

\subsection{Experiments on Real Data}
We also test our approach on the OAEI\footnote{The Ontology Alignment Evaluation Initiative (OAEI) provides reference alignments and carries out yearly evaluation campaigns since 2004. For more information, see \url{http://oaei.ontologymatching.org/}} multifarm dataset. Here, multilingual ontologies from the conference domain have to be matched. Since the absolute orientation approach does not use textual data, we only evaluate the German-English test case. This is sufficient because the other language combinations of the multifarm dataset use structurally identical graphs.
With a sampling rate of 20\%, our approach achieves micro scores of $P=0.376$, $R=0.347$, and $F_1=0.361$. Compared to the systems participating in the 2021 campaign~\cite{DBLP:conf/semweb/PourAAAFFFHHHHI21}, the recall is on par with state of the art systems; an overall lower $F_1$ is caused by a comparatively low precision score.
While not outperforming top-notch OAEI systems in terms of $F_1$, the performance indicates that the approach is able to perform ontology matching and may particularly benefit from the addition of non-structure-based features.

\section{Conclusion}
In this paper, we showed early work on aligning graphs through a graph embedding algorithm combined with an absolute orientation rotation approach. In multiple experiments we showed that the approach works for structurally similar ontologies. It handles alignment noise better than varying sizes and structures of graphs. In the future, we plan to conduct experiments with different variants of embedding approaches \cite{DBLP:conf/semweb/PortischP21,portisch2022walk}, as well as to combine the approach with further text-based features in a hybrid matching system.

\bibliographystyle{splncs04}
\bibliography{references}

\end{document}